%% file: main.tex
\documentclass[10pt,twocolumn,letterpaper]{article}

\usepackage[pagenumbers]{cvpr} % To force page numbers, e.g. for an arXiv version

\usepackage{multirow}
\usepackage{bbding}
\usepackage{overpic}
\usepackage[table,cmyk]{xcolor}
\usepackage{color}

\definecolor{myRed}{rgb}{0.8,0.2,0.2}
\definecolor{myBlue}{rgb}{0.2,0.2,0.8}

\definecolor{cvprblue}{rgb}{0.21,0.49,0.74}
\usepackage[pagebackref,breaklinks,colorlinks,allcolors=cvprblue]{hyperref}

\def\paperID{35954} % *** Enter the Paper ID here
\def\confName{CVPR}
\def\confYear{2026}

\makeatletter
\DeclareRobustCommand\onedot{\futurelet\@let@token\@onedot}
\def\@onedot{\ifx\@let@token.\else.\null\fi\xspace}

\def\eg{\emph{e.g}\onedot}

\def\etal{\emph{et al}\onedot}

\makeatother

\newcommand{\figref}[1]{Fig.~\ref{#1}}
\newcommand{\tabref}[1]{Tab.~\ref{#1}}

\newcommand{\secref}[1]{Sec.~\ref{#1}}

\title{3D-Fixer: Coarse-to-Fine In-place Completion for 3D Scenes \\from a Single Image}

\author{Ze-Xin Yin\renewcommand{\thefootnote}{\arabic{footnote}}\footnotemark[1]\ ~\renewcommand{\thefootnote}{\fnsymbol{footnote}}\footnotemark[1]
\qquad Liu Liu\renewcommand{\thefootnote}{\arabic{footnote}}\footnotemark[3]\ ~\renewcommand{\thefootnote}{\fnsymbol{footnote}}\footnotemark[2]
\qquad Xinjie Wang\renewcommand{\thefootnote}{\arabic{footnote}}\footnotemark[3]
\qquad Wei Sui\renewcommand{\thefootnote}{\arabic{footnote}}\footnotemark[4]
\qquad Zhizhong Su\renewcommand{\thefootnote}{\arabic{footnote}}\footnotemark[3]
\qquad Jian Yang\renewcommand{\thefootnote}{\arabic{footnote}}\footnotemark[1]  
\qquad Jin Xie\renewcommand{\thefootnote}{\arabic{footnote}}\footnotemark[2]\ ~\renewcommand{\thefootnote}{\fnsymbol{footnote}}\footnotemark[3]\\
\renewcommand{\thefootnote}{\arabic{footnote}}\footnotemark[1] \ College of Computer Science, Nankai University \\
\renewcommand{\thefootnote}{\arabic{footnote}}\footnotemark[2] \ School of Intelligence Science and Technology, Nanjing University \\
\renewcommand{\thefootnote}{\arabic{footnote}}\footnotemark[3] \ Horizon Robotics \qquad
\renewcommand{\thefootnote}{\arabic{footnote}}\footnotemark[4] \ D-Robotics \\
}

\begin{document}

\twocolumn[
    \maketitle
    \vspace{-2.8em}
    \input{sec/teaser}

    \bigbreak
]

\input{sec/commands}

\input{sec/0_abstract}    
\renewcommand{\thefootnote}{\fnsymbol{footnote}}
\footnotetext[1]{Intern at D-Robotics: \tt\small zexin.yin.cn@mail.nankai.edu.cn}
\renewcommand{\thefootnote}{\fnsymbol{footnote}}
\footnotetext[2]{Project leader.}
\footnotetext[3]{Corresponding author: \tt\small csjxie@nju.edu.cn}
\input{sec/1_intro}
\input{sec/2_related_works}

\input{sec/3_methods}
\input{sec/4_dataset}
\input{sec/5_exp}

\input{sec/6_conclusion}

{
    \small
    \bibliographystyle{ieeenat_fullname}
    \bibliography{main}
}

\input{sec/X_suppl}

\end{document}

%% file: sec/teaser.tex
\begin{center}
\vspace{6pt}
\centering
\begin{overpic}[width=\linewidth]{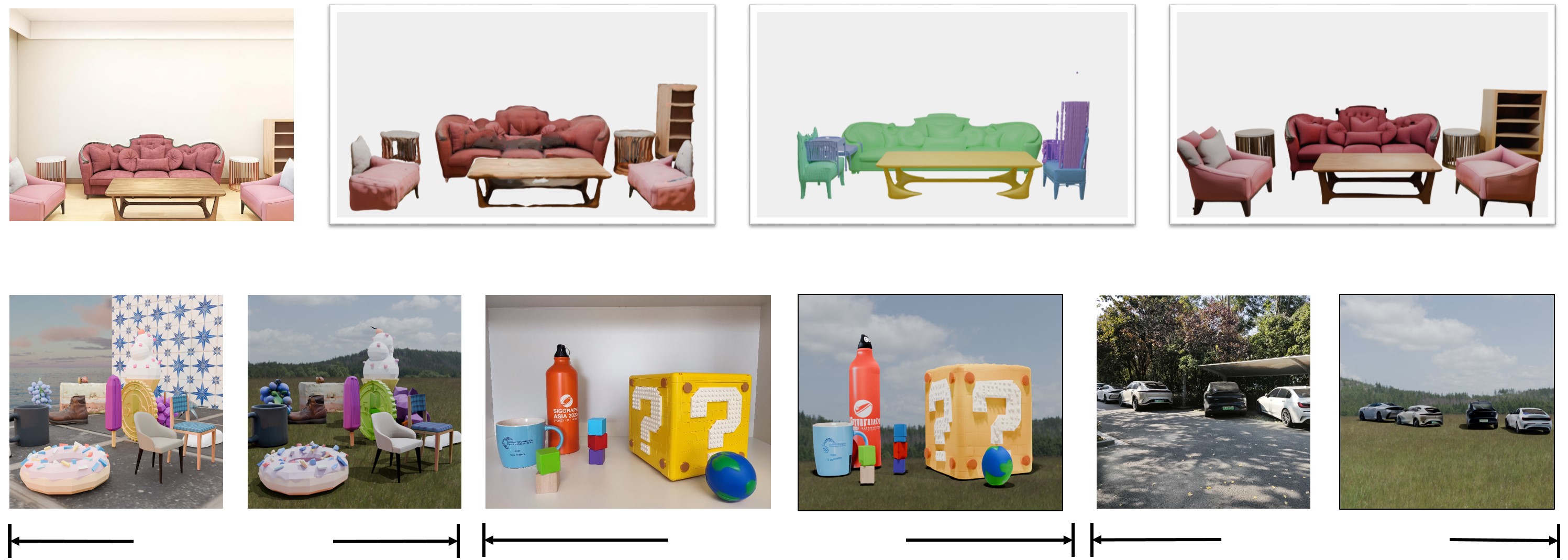}
\put(41, 36){(a) Scene generation}
\put(4, 19.5){Input image}
\put(29.5, 19.5){Gen3DSR}
\put(58, 19.5){MIDI}
\put(85.5, 19.5){Ours}

\put(42, 18){(b) Generalization}
\put(8.7, 0.8){Complex scene}
\put(43.2, 0.8){Real world scene}
\put(79.5, 0.8){Outdoor scene}
\end{overpic}
\vspace{-10pt}
\captionof{figure}{
Performance overview. 3D-Fixer extends pre-trained image-to-3D generative priors to 
achieve compositional 3D scene generation through a novel in-place completion paradigm.
(a) Our method significantly outperforms baselines such as Gen3DSR and MIDI in geometry quality. (b) It further demonstrates strong generalization to complex real-world and outdoor scenes.
}
\vspace{-6pt}
\label{fig:teaser}
\end{center}

%% file: sec/commands.tex
\newcommand{\myPara}[1]{\vspace{.05in}\noindent\textbf{#1.}}

%% define color for table
\definecolor{best}{hsb}{0.667,0.6,1}
\definecolor{second}{hsb}{0.583,0.6,1}
\definecolor{third}{hsb}{0.5,0.6,1}
\definecolor{default}{hsb}{0,0,1}

%% file: sec/0_abstract.tex
\begin{abstract}

Compositional 3D scene generation from a single view requires the simultaneous recovery of scene layout and 3D assets.
Existing approaches mainly fall into two categories: feed-forward generation methods and per-instance generation methods.
The former directly predict 3D assets with explicit 6DoF poses through efficient network inference, but they generalize poorly to complex scenes.
The latter improve generalization through a divide-and-conquer strategy, but suffer from time-consuming pose optimization. 
To bridge this gap, we introduce \textbf{3D-Fixer}, a novel in-place completion paradigm.
Specifically, \textbf{3D-Fixer} extends 3D object generative priors to generate complete 3D assets conditioned on the partially visible point cloud at the original locations,
which are cropped from the fragmented geometry obtained from the geometry estimation methods.
Unlike prior works that require explicit pose alignment, 
\textbf{3D-Fixer} uses fragmented geometry as a spatial anchor to preserve layout fidelity.
At its core, we propose a coarse-to-fine generation scheme to resolve boundary ambiguity under occlusion, 
supported by a dual-branch conditioning network and an Occlusion-Robust Feature Alignment (ORFA) strategy for stable training.
Furthermore, to address the data scarcity bottleneck, we present ARSG-110K, the largest scene-level dataset to date, 
comprising over 110K diverse scenes and 3M annotated images with high-fidelity 3D ground truth. 
Extensive experiments show that 3D-Fixer achieves state-of-the-art geometric accuracy,
which significantly outperforms baselines such as MIDI and Gen3DSR, 
while maintaining the efficiency of the diffusion process. Code and data will be publicly available at \href{https://zx-yin.github.io/3dfixer}{https://zx-yin.github.io/3dfixer}.

\end{abstract}

%% file: sec/1_intro.tex
\section{Introduction}
\label{sec:intro}

Single view compositional 3D scene reconstruction is a challenging yet crucial task for various applications, including robotics, embodied AI, AR/VR, \etal.
This task requires inferring geometry, texture, and spatial layout simultaneously from limited visual information.
Recently, there has been a growing trend toward incorporating strong 3D visual priors~\cite{huang2025midi,ardelean2025gen3dsr,meng2025scenegen}, 
including object-level generative priors~\cite{xiang2025structured} and geometry foundation models~\cite{wang2025moge,wang2025vggt}, 
to tackle this challenging task.

While recent advances have been significant, current methods still struggle to balance efficiency, fidelity, and generalization.
We categorize existing approaches into two classes: feed-forward scene generation methods and divide-and-conquer methods.
The former methods~\cite{huang2025midi,meng2025scenegen,lin2025partcrafter,chen2024single,chu2023buol,dahnert2021panoptic,gkioxari2022learning,liu2022instpifu,nie2020total3dunderstanding,paschalidou2021atiss,zhang2021holistic,zhang2023uni} feature end-to-end networks that take a scene image as input and predict 3D assets aligned with the scene layout via network inference,
but require massive high-quality scene-level training data.
Despite their efficiency, the scarcity of scene-level training data severely limits their generalization to open-set real-world scenarios.
The latter methods~\cite{ardelean2025gen3dsr, yao2025cast,gao2024diffcad, gumeli2022roca,izadinia2017im2cad, kuo2020mask2cad, kuo2021patch2cad, langer2022sparc, yan2023psdr} 
divide the scene into individual objects, retrieve or generate each 3D asset, and align them with the observations via an optimization process.
Although the divide-and-conquer strategy improves generalization, retrieval or generation in real scenes with occlusions introduces misalignment between the 3D assets and the actual objects,
and the pose optimization process is time-consuming and error-prone. 

Existing divide-and-conquer methods improve generalization by leveraging estimated geometry as a strong spatial anchor and aligning 
3D assets to the scene via 2D reprojection~\cite{ardelean2025gen3dsr} or 3D point cloud registration~\cite{yao2025cast}.
Although recent geometry foundation models~\cite{wang2025moge, wang2025vggt, bochkovskii2024depth, yang2024depth,yang2024depthv1, sun2025depth, yin2023metric3d, wang2025mogev1, wang2024dust3r, wang2025pi}
have made substantial progress in accurately recovering the spatial layout of complex real-world scenes,
the alignment process can still fail due to errors accumulated during 3D asset acquisition
since the retrieved or generated 3D assets may be inconsistent with the scene instances in single-view settings.
Since this alignment process is inherently error-prone, we ask whether layout priors can be better exploited without relying on explicit alignment.
By revisiting the estimated geometry, we find that it contains not only the spatial layout but also the visible parts of each scene instance.
This observation inspires us to \textit{complete the unseen parts of each instance from its visible geometry while preserving its original location},
so that the scene can be generated naturally.
In this paper, we explore this novel in-place completion paradigm, as illustrated in \figref{fig:pip}.

Built on this in-place completion paradigm, we present \textit{3D-Fixer}, 
a novel framework that adapts a pre-trained object generation model~\cite{xiang2025structured} to perform simultaneous multi-instance completion 
conditioned on monocular geometry estimation for compositional 3D scene generation.
To effectively adapt object-level generative priors to the in-place completion task,
3D-Fixer adopts a dual-branch conditioning scheme inspired by \cite{zhang2023adding}, in which an additional branch processes scene-level 2D and 3D contextual cues while leaving the original object generation branch unchanged.
In real-world scenes, in-place completion suffers from boundary ambiguity caused by occlusion and limited viewpoint observations.
To address this issue, we employ a coarse-to-fine strategy, where 3D-Fixer first estimates a loose boundary to capture the global structure, 
then refines each asset within a tightened spatial bound.
Moreover, object-level generative priors are trained on clean, occlusion-free images, 
whereas occlusions are common in scene-level settings.
We therefore introduce an Occlusion-Robust Feature Alignment (ORFA) strategy to stabilize training.
By aligning features with those of a frozen clean-input object branch, 3D-Fixer is guided to maintain structural plausibility in complex scenarios.

Beyond the framework, we also address the data scarcity bottleneck in this line of research.
Existing datasets~\cite{fu20213dfront, fu20213dfuture, yu2025metascenes} often suffer from limited scale, low diversity, or a lack of object-level ground truth.
To overcome these limitations, we introduce Asset-Rich Scene Generation (ARSG-110K), a large-scale dataset for compositional 3D scene generation.
Leveraging the Blender Cycles engine~\cite{blender}, we procedurally generate over 110K scenes using a library of more than 180K high-quality assets, 1K HDR maps, and 5K textures, 
yielding over 3M valid views.
Each scene features complex compositions of up to 20 assets and comprehensive annotations, including camera parameters, instance masks, and accurate object-level geometry with 6DoF poses.
This resource not only supports the training of 3D-Fixer but also establishes a rigorous benchmark for future research.

In summary, our key contributions are as follows:

\begin{itemize}
    \item 
    We propose \textit{3D-Fixer}, a novel framework for single-view compositional 3D scene generation via an in-place completion scheme.
    By combining the generalization capabilities of pre-trained object-level generative models with the structural fidelity of 3D geometry foundation models, 
    3D-Fixer achieves state-of-the-art performance while maintaining the efficiency of feed-forward diffusion.

    \item 
    We design a dual-branch conditioning mechanism to effectively adapt object-level generative priors for scene generation,
    together with an Occlusion-Robust Feature Alignment (ORFA) training strategy and a coarse-to-fine generation scheme.
    This design resolves boundary ambiguity and enables robust completion in real-world scenes, offering clear advantages over existing methods.

    \item 
    We build and release ARSG-110K (Asset-Rich Scene Generation), a large-scale dataset designed to address the data scarcity bottleneck.
    To our knowledge, it is the largest open-source dataset for compositional 3D scene generation. 
    Comprising over 110,000 scenes with high-quality object-level ground truth and detailed annotations, it facilitates future research in this area.
\end{itemize}

%-------------------------------------------------------------------------

%% file: sec/2_related_works.tex
\section{Related Works}
\label{sec:rel_work}

\begin{figure*}
  \centering
  \vspace{-12pt}
  \begin{subfigure}{0.95\linewidth}
    \includegraphics[width=\linewidth]{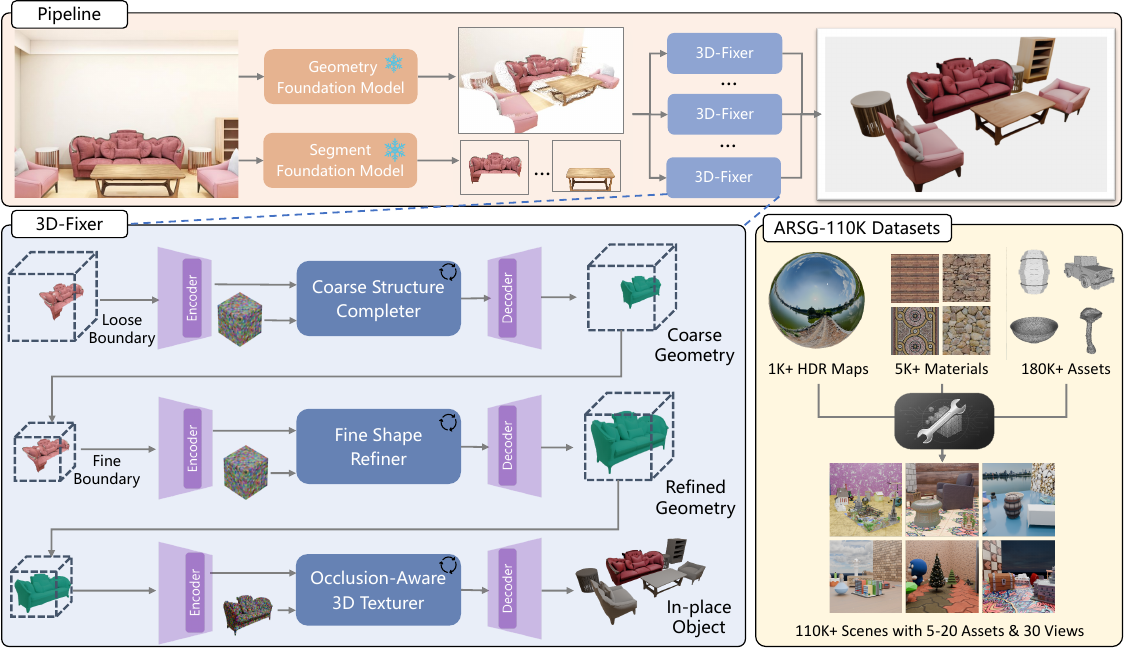}
  \end{subfigure}
  \caption{Architecture of the 3D-Fixer pipeline and dataset. (Top) \textit{Scene Decomposition} extracts instance-level partial geometry from the input. (Bottom-left) \textit{Progressive Completion} generates the full asset via three stages: 1) The \textit{Coarse Structure Completer} hallucinates topology within a loose bound; 2) The \textit{Fine Shape Refiner} sharpens geometry within a fine boundary; and 3) The \textit{Occlusion-Aware 3D Texturer} applies observation-aligned textures. (Bottom-right) Our \textit{ARSG-110K Dataset} provides high-quality assets and rich scene compositions for training.}
  
\label{fig:pip}
\vspace{-6pt}

\end{figure*}

Existing single view compositional scene generation methods can be broadly classified into two categories:
feed-forward generation methods and per-instance retrieval or generation methods.
This section reviews these lines of work to highlight how our approach differs from existing methods.
In addition, recent developments in geometry foundation models and object-level generation methods inspire our method and form its foundation.
Therefore, we also summarize them in this section.

\subsection{Compositional Scene Generation}
\myPara{Feed-Forward Generation Methods}
Given a scene image and multiple instance masks, these methods reconstruct multiple 3D assets in the scene via feed-forward inference or a diffusion process.
Diffusion-based methods~\cite{huang2025midi,lin2025partcrafter,meng2025scenegen} achieve high-quality component generation for indoor scenes.
Early methods~\cite{dahnert2021panoptic, nie2020total3dunderstanding, liu2022instpifu,chen2024single} enable efficient layout and geometry generation via feed-forward inference.
However, due to limitations in dataset scale and diversity for this task, existing methods struggle to generalize to complex real-world scenes.
Furthermore, methods such as MIDI~\cite{huang2025midi} and SceneGen~\cite{meng2025scenegen} introduce multi-instance attention, 
leading to computational complexity that scales quadratically with the number of objects in a scene, 
which severely limits their scalability in environments containing many 3D instances.

\myPara{Per-instance Generation and Optimization Methods}
To overcome the limited generalization of feed-forward generation, some methods~\cite{yao2025cast,ardelean2025gen3dsr,han2025reparo, gao2024diffcad, yan2023psdr, han2025reparo} 
follow a divide-and-conquer strategy that
decomposes the task into 3D object generation~\cite{xiang2025structured, lin2025partcrafter, li2025step1x, 
wu2024direct3d, zhang20233dshape2vecset,lai2025hunyuan3d25highfidelity3d,hunyuan3d22025tencent,yang2024hunyuan3d} or 3D object  
retrieval from existing 3D asset dataset~\cite{deitke2023objaverse, deitke2023objaverse, lin2025objaversepp, collins2022abo, chang2015shapenet}, followed by pose alignment.
These methods leverage powerful pre-trained object-level generative models to generate each scene instance individually,
and then iteratively optimize the pose and scale of each complete 3D asset to align it with the input view.
However, the iterative procedure increases computational cost,
and the optimization is prone to local minima and accumulated registration errors,
limiting robustness. 
In contrast, our method achieves comparable completeness and generalization without sacrificing efficiency, 
by performing in-place completion directly in the scene and thereby avoiding an error-prone alignment.

\myPara{Feed-forward Geometry Estimation Methods}
Unlike 3D generation methods, geometry estimation methods focus on direct 3D reconstruction from single or multiple views.
These methods predict various representations, such as depth maps~\cite{wang2025moge,yang2024depth,bochkovskii2024depth},
and point clouds~\cite{wang2025vggt,wang2025pi,deng2025vggtlongchunkitloop}, achieving high-quality geometry estimation for visible regions.
However, they recover only the geometry observed in the input view(s).
As a result, the estimated geometry is fragmented and lacks closed boundaries and occluded structures.
Despite this limitation, these methods provide accurate local pose and layout information for the scene.
Our 3D-Fixer introduces a novel in-place completion paradigm that leverages the layout information from geometry estimation methods 
as an initial geometric constraint and uses 3D generative priors to produce complete 3D assets.

\subsection{Image-based Object-Level Generation}
The field of object-level 3D generation from a single image has seen rapid development, driven 
by powerful generative models\cite{ho2020denoising,lipman2023flowmatchinggenerativemodeling}. 
Existing methods, including \cite{xiang2025structured, lin2025partcrafter, li2025step1x, 
wu2024direct3d, zhang20233dshape2vecset,lai2025hunyuan3d25highfidelity3d,hunyuan3d22025tencent,yang2024hunyuan3d},
typically adopt a two-stage generation process.
They first generate coarse object geometry in the form of a point cloud or voxel grid,
and then refine the details and decode the 3D models into representations such as SDFs or radiance fields.
Our method is built upon the TRELLIS~\cite{xiang2025structured} architecture and its SLAT representation,
adapting pre-trained object-level 3D generative priors to our in-place completion paradigm.

%% file: sec/3_methods.tex
\section{Methods}
\label{sec:method}

We propose 3D-Fixer, a novel and robust framework for single-image compositional 3D scene generation by extending object-level generative priors.
Our method introduces a novel in-place completion paradigm by combining 3D generative priors with geometry estimation methods.
This design eliminates the need for error-prone pose alignment while preserving the high efficiency of feed-forward diffusion inference.

% \subsection{Preliminaries: Object-Level Generative Priors}
% \label{subsec:prel}

\myPara{Preliminaries: Object-Level Generative Priors}
Our framework adapts TRELLIS~\cite{xiang2025structured} as our foundational object generation prior, a two-stage model using flow matching~\cite{lipman2023flowmatchinggenerativemodeling} and a unified Structured LATent (SLAT) representation. The first stage generates a coarse representation by training a DiT on a latent space, which is compressed from a $64^3$ volumetric grid by a 3D Voxel VAE. The second stage then operates on these coarse voxels, where a sparse VAE compresses aggregated image features (\eg, from DINOv2~\cite{oquab2023dinov2}), and a sparse DiT generates high-fidelity latent features that are decoded into the final mesh.

However, this foundational prior is designed for single, isolated objects and cannot natively handle the completion of occluded instances or the compositional layout recovery required in a full scene context.

\begin{figure}
  \centering
  \includegraphics[width=\linewidth]{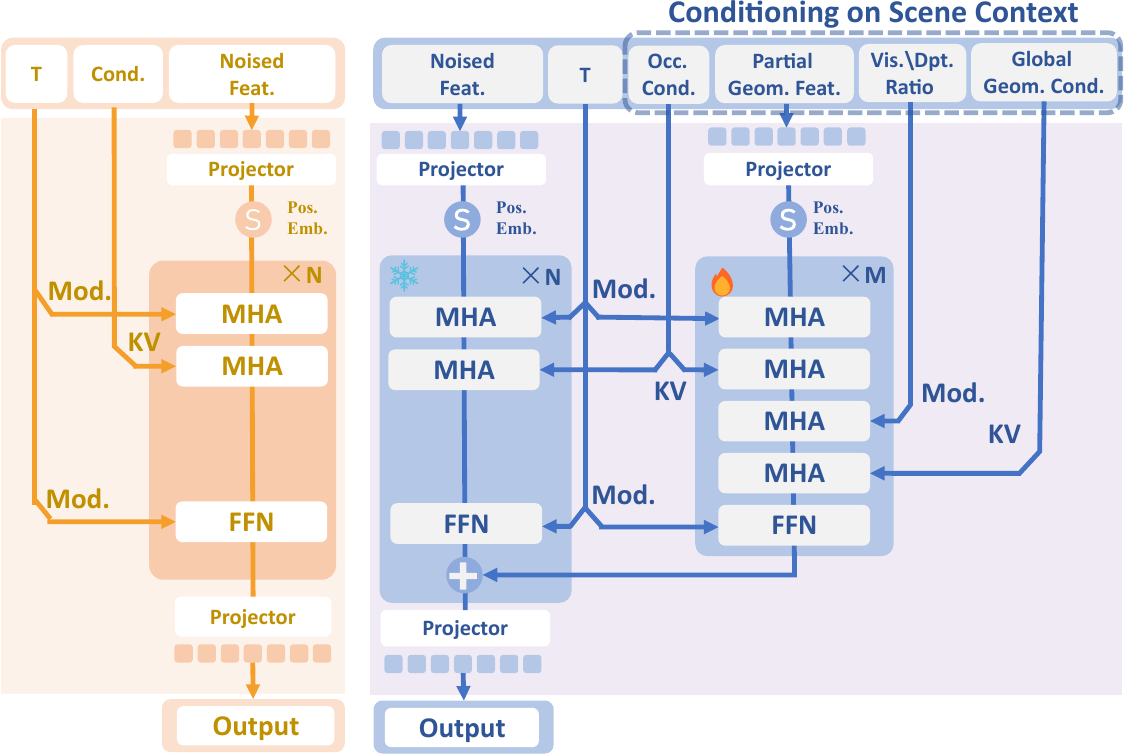}
  \caption{3D-Fixer extends the diffusion transformer from prior~\cite{xiang2025structured} (orange) to a dual-stream architecture (blue), where a trainable branch encoding scene-specific geometric cues interacts with a frozen generative branch to implement ORFA and enforce structural constraints.}
    \label{fig:architecture}
    \vspace{-12pt}
\end{figure}

\subsection{3D-Fixer Pipeline Overview}
\label{subsec:inplace_gen}

The core of our method is the 3D in-place completion paradigm, which achieves geometry and layout fidelity in an efficient manner.
Unlike existing methods requiring canonical point cloud conditions~\cite{yao2025cast}, our approach conditions the generative prior directly on the scene's spatial context.
As illustrated in Figure~\ref{fig:pip}, our pipeline first performs scene decomposition: given an image $I$, it yields instance masks $\{M_i\}$~\cite{ravi2024sam, ren2024grounded} and fragmented point clouds $G_{\mathrm{frag}}$~\cite{wang2025moge, wang2025vggt}. 
This fragmented geometry serves as the geometric condition for our progressive completion phase, which is applied in parallel to each instance. 
This phase first generates the overall topology within a loose bound to handle boundary uncertainty (Coarse Structure Completer), then sharpens geometric details using the predicted fine boundary (Fine Shape Refiner), and finally applies photorealistic textures aligned with the 2D view (Occlusion-Aware 3D Texturer).

Our key innovations focus on three technical contributions: (1) a Contextual Conditioning mechanism (Sec.~\ref{subsec:net}) that robustly integrates 2D and 3D cues, including our Geometry-Aware Feature Projection (GAFP); (2) a Coarse-to-Fine Generation scheme (Sec.~\ref{subsec:c2f}) to accurately resolve boundary uncertainty from occlusion; and (3) an Occlusion-Robust Training strategy (Sec.~\ref{subsec:train}) that leverages teacher guidance via an ORFA loss to stabilize training.

\subsection{Conditioning on Scene Context}
\label{subsec:net}

\myPara{Geometry Contextual Conditioning}
Existing feed-forward generation methods often rely solely on 2D information (e.g., image $I$ and masks $\{M_i\}$) as conditions. 
This neglects the crucial 3D spatial information of an object's visible part, 
leading to significant geometric ambiguity in scale and orientation.
To overcome this, 3D-Fixer explicitly incorporates the visible, fragmented point cloud $G_{\mathrm{frag}}$ and its corresponding mask as a geometric condition.
This explicit 3D information acts as a strong spatial anchor, grounding the generative process and guiding the prior to synthesize a complete shape that is precisely aligned with the visible geometry.
Since the fragmented point cloud $G_{\mathrm{frag}}$ inevitably contains distortions in complex scenes,
3D-Fixer designs the conditioning branch with depth-ratio-embedded self-attention and global-feature cross-attention to handle varying degrees of distortion.
We provide further analysis in the Supplementary.

\myPara{Texture Contextual Conditioning}
While the base prior~\cite{xiang2025structured} utilizes DINOv2~\cite{oquab2023dinov2} tokens via cross-attention, this mechanism only establishes a weak, global correspondence between the image and the 3D geometry. It lacks the precise spatial grounding needed for high-fidelity texture synthesis under occlusion. To establish a stronger, localized correspondence, we introduce a Geometry-Aware Feature Projection (GAFP) mechanism. As shown in Figure~\ref{fig:architecture}, GAFP projects high-resolution 2D image features from MoGe v2~\cite{wang2025moge} directly onto the 3D voxel coordinates of the visible point cloud $G_{\mathrm{frag}}$. 
This projection creates a set of spatially-aligned appearance features that act as a powerful, localized condition. These projected features are processed by a conditional branch, which further enriches the information using visibility-ratio-embedded self-attention and injecting global-feature tokens from MoGe v2~\cite{wang2025moge} via cross-attention. These features are then injected layer-wisely into the DiT blocks to guide texture generation.

\subsection{Coarse-to-Fine Generation Scheme}
\label{subsec:c2f}

Occlusion is pervasive in 3D scenes, often causing the visible point cloud $G_{\mathrm{frag}}$ to be significantly smaller than the complete asset $A$. This creates severe boundary ambiguity for in-place completion. To handle this, we propose an efficient coarse-to-fine scheme that decouples boundary prediction from detail generation.

\myPara{Coarse Contour Prediction}
We first compute the minimum axis-aligned bounding box (AABB) $B_{\mathrm{vis}}$ of the input fragmented point cloud $G_{\mathrm{frag}}$. 
We then define a conservative, expanded bounding box $B_{\mathrm{exp}}$ centered at $B_{\mathrm{vis}}$'s center, $C_{\mathrm{vis}}$, with a side length $4\times$ the maximum side length of $B_{\mathrm{vis}}$. 
This conservative expansion ensures that the unknown complete boundary $B_{\mathrm{full}}$ is contained within $B_{\mathrm{exp}}$ even under heavy occlusion.
The model first performs a coarse in-place completion within this expanded volume $B_{\mathrm{exp}}$. 
The sole objective of this stage is to predict the accurate complete object boundary $B_{\mathrm{full}}$, as shown in Figure~\ref{fig:pip}.

\myPara{Fine Geometry Generation}
With the predicted boundary $B_{\mathrm{full}}$, we now have a precisely defined volume. 
In this stage, we fully leverage the pre-trained geometric prior to generate the final, high-resolution, and high-fidelity geometry within this boundary. 
This two-stage strategy successfully decouples scale prediction from geometry generation: the coarse stage resolves boundary uncertainty, while the fine stage focuses on high-fidelity completion, ensuring a geometrically accurate and complete asset.

\subsection{Occlusion-Robust Training}
\label{subsec:train}

The base prior~\cite{xiang2025structured} is pre-trained on large-scale datasets of clean, unoccluded single-object views. Adapting this prior to our in-place completion task introduces a significant domain gap due to scene-level occlusion, which can cause training instability. 
Specifically, severe occlusion creates a one-to-many mapping ambiguity between the limited visible point cloud ($G_{\mathrm{frag}}$) and the final complete asset ($A$), hindering stable convergence.

To stabilize training, we introduce an Occlusion-Robust Feature Alignment (ORFA) strategy. 
Inspired by~\cite{yu2024representation, wu2025representation},
ORFA employs knowledge distillation, using the frozen, pre-trained TRELLIS model as a teacher to guide our scene-conditioned model in a layer-wise manner.

During training, in addition to the standard Flow Matching loss ($L_{\text{FM}}$), we impose an Alignment Loss ($L_{\text{AL}}$) that utilizes a frozen teacher model to constrain the 3D-Fixer. Specifically, both the teacher and 3D-Fixer process the same noised feature $\mathbf{z}_t$ at noise level $t$. However, the teacher is conditioned on the clean image to produce intermediate latent representations $\{\mathbf{h}\}$, whereas 3D-Fixer is conditioned on the occluded input to generate representations $\{\mathbf{h}_{s}\}$. The alignment loss is defined as follows:

\begin{equation}
    \mathcal{L}_{\text{AL}} := -\mathbb{E}\Big[ \frac{1}{N}\sum_{n=1}^{N} \mathrm{sim}(\mathbf{h}_{s}, \mathbf{h}) \Big]
\end{equation}

This alignment loss acts as a powerful regularizer. 
The dual constraint mitigates the detrimental effects of occlusion-induced ambiguity. 
It ensures the generative prior retains its strong, unoccluded shape knowledge while simultaneously learning to incorporate the new scene-level contextual cues, leading to stable adaptation for completion in complex, occluded scenarios.

%% file: sec/4_dataset.tex
\section{Dataset for Scene Generation}
\label{sec:data}

\begin{table}[ht]
    \vspace{-6pt}
    \centering
    \caption{We compare ARSG-110K with existing 3D scene datasets. Our dataset significantly scales up scene and object diversity through procedural generation, offering over 3 million rendered views.}
    \label{tab:comparison}
    
    \resizebox{0.48\textwidth}{!}{%
        \begin{tabular}{lcccc}
            \toprule
            \toprule
            Dataset & Source & Objects & Scenes & Images \\
            \midrule
            \midrule
            Scan2CAD~\cite{avetisyan2019scan2cad} & ShapeNet~\cite{chang2015shapenet} & 14.2K & 706 & 2.5M \\
            OpenRooms~\cite{li2021openrooms} & ShapeNet~\cite{chang2015shapenet} & 16.0K & 706 & 2.5M \\
            METASCENES~\cite{yu2025metascenes} & ScanNet~\cite{dai2017scannet}, Objaverse~\cite{deitke2023objaverse} & 15.4K & 706 & 2.5M \\
            R3DS~\cite{wu2024r3ds} & ShapeNet~\cite{chang2015shapenet}, Wayfair~\cite{sadalgi2016} & 19.1K & 370 & 194K \\
            CAD-Estate~\cite{maninis2023cad} & ShapeNet~\cite{chang2015shapenet} & 100K & 19.5K & - \\
            BVS~\cite{ge2024behavior} & BEHAVIOR-1K~\cite{li2023behavior} & 6.7K & 1000 & - \\
            HSSD-200~\cite{wu2024r3ds} & Floorplanner & 18.7K & 211 & - \\
            3D-FRONT~\cite{fu20213dfront} & 3D-FUTURE~\cite{fu20213dfuture} & 13.2K & 18.9K & 20K \\
            \textbf{ARSG-110K (Ours)} & 
            \begin{tabular}{@{}c@{}}
            Objaverse~\cite{lin2025objaversepp}, 3D-FUTURE~\cite{fu20213dfuture}, \\ 
            HSSD~\cite{khanna2024habitat}, ABO~\cite{collins2022abo}
            \end{tabular}
            & \textbf{180K+} & \textbf{110K} & \textbf{3M} \\
            \bottomrule
            \bottomrule
        \end{tabular}%
    }
    \vspace{-12pt}
\end{table} 

% %% 
As summarized in \tabref{tab:comparison},
existing scene-level datasets provide context but are bottlenecked by ground-truth (GT) quality or scale. 
3D-Front includes complete furniture assets but is small.
Large-scale real indoor datasets like ScanNet and Matterport3D offer RGB-D data but lack true object-level 3D assets. 
The MetaScenes~\cite{yu2025metascenes} method proposes a strategy to generate 3D asset GT for ScanNet scenes, but the resulting scene still suffers from minor misalignments or distortions compared to the original scene. 

To address the scarcity of training data that contains complex scene layouts with high-fidelity object geometry, we introduce ARSG-110K. Unlike existing datasets that often sacrifice object quality for scene scale, our dataset preserves accurate instance-level ground truth within diverse, procedurally generated environments. This unique combination establishes a rigorous benchmark for both object- and scene-level 3D generation research.

To ensure photorealistic visual quality, we employ the Blender Cycles renderer backed by a massive repository of resources. Specifically, our construction pipeline leverages over 180K high-quality 3D assets across diverse categories, 1K+ HDR maps for realistic environmental lighting, and 5K+ material textures to diversify scene boundaries such as floors and walls. Using an automated procedural script, we constructed over 110K unique scene configurations. Each scene is densely populated with 5--20 individual assets, purposefully creating complex object-to-object occlusion scenarios to challenge and train robust generative models.

The resulting dataset offers a massive scale of high-quality training data. We rendered 30 random camera views for each scene, yielding over 3 million images. Crucially, each rendered view is paired with comprehensive annotations, including camera intrinsics and extrinsics, per-pixel object instance masks, and complete 3D mesh models for every asset, accompanied by their precise translation and rotation matrices in the scene coordinate.

%% file: sec/5_exp.tex
\section{Experiments}
\label{sec:exp}

In this section, we present comprehensive experimental validation of our proposed 3D-Fixer framework. 
We first detail the experimental setup, including the datasets, evaluation metrics, and baselines.
The training details are in the Supplementary.
We then provide quantitative comparisons against state-of-the-art methods and conduct a 
thorough ablation study to analyze the effectiveness of our key architectural components.

\subsection{Hyperparameters and Benchmarks}
\label{subsec:setup}

\myPara{Evaluation}
We train 3D-Fixer on our synthetic dataset, and test the performance across multiple existing scene generation benchmarks,
including MIDI testset and Gen3DSR testset. %  and real-world images from Gen3DSR.
In addition to the existing dataset, we further select 3D assets from the Toys4K~\cite{stojanov21toys4k} data set and construct 100 random scenes 
following the same procedure as the ARSG-110K dataset, serving as our new testset.
Furthermore, we select 15 scenes from ScanNet with MetaScenes~\cite{yu2025metascenes} serving as the ground-truth to evaluate the performance
on real-world scenes.
Following existing methods~\cite{huang2025midi, ardelean2025gen3dsr}, we report scene-level Chamfer Distance and F-Score 
with the threshold of 0.1 on the compositional scenes.
We also report object-level Chamfer Distance and F-Score for each object in the scene.
Moreover, we leverage the Volumetric Intersection over Union between the bounding boxes of the generated objects
and the ground truth in the scenes.
Due to the organization of these dataset, we report scene-level metrics on all evaluation benchmarks,
but only report object-level metrics on our testset and the MIDI testset.
To demonstrate efficiency, we report the inference time of our network on the MIDI testset using an NVIDIA RTX 5090 GPU.

\myPara{Baselines}
We mainly compare our method with the state-of-the-art diffusion method MIDI~\cite{huang2025midi},
and the per-instance generation and optimization method Gen3DSR~\cite{ardelean2025gen3dsr}.
In addition, we report metrics for early feed-forward methods PanoRecon~\cite{dahnert2021panoptic}, Total3D~\cite{nie2020total3dunderstanding},
InstPIFu~\cite{liu2022instpifu}, and SSR~\cite{chen2024single}, retrieval-based methods DiffCAD~\cite{gao2024diffcad},
and per-instance optimization methods REPARO~\cite{han2025reparo}.

\begin{table}[t]
  \centering
  \vspace{-5pt}
  \caption{Quantitative comparisons on synthetic datasets~\cite{fu20213dfront,ardelean2025gen3dsr,chen2024single} and real-world dataset~\cite{dai2017scannet,yu2025metascenes}. We report Scene ($S$) and Object ($O$) level Chamfer Distance (CD) and F-Score (FS), along with Bounding Box IoU and inference time.}
  \vspace{-8pt}
  \begin{subtable}[t]{\linewidth}
  \caption{Results on MIDI testset.}
    \label{tab:midi_res}
    \resizebox{\linewidth}{!}{
      \begin{tabular}{ccccccc}
        \toprule
        \toprule
           Method & CD$_S$$\downarrow$ & FS$_S$$\uparrow$ & CD$_O$$\downarrow$ & FS$_O$$\uparrow$ & IoU$\uparrow$ & Time$\downarrow$ \\
        \midrule
        \midrule
            PanoRecon~\cite{dahnert2021panoptic} & 0.150 & 40.65 & 0.211 & 35.05 & 0.240 & 32s \\
            Total3D~\cite{nie2020total3dunderstanding} & 0.270 & 32.90 & 0.179 & 36.38 & 0.238 & 39s \\
            InstPIFu~\cite{liu2022instpifu} & 0.138 & 39.99 & 0.165 & 38.11 & 0.299 & 32s \\
            SSR~\cite{chen2024single} & 0.140 & 39.76 & 0.170 & 37.79 & 0.311 & 32s \\
            DiffCAD~\cite{gao2024diffcad} & 0.117 & 43.58 & 0.190 & 37.45 & 0.392 & 64s \\
            Gen3DSR~\cite{ardelean2025gen3dsr} & 0.123 & 40.07 & 0.157 & 38.11 & 0.363 & 9m \\
            REPARO~\cite{han2025reparo} & 0.129 & 41.68 & 0.160 & 40.85 & 0.339 & 4m \\
            MIDI~\cite{huang2025midi} & 0.080 & 50.19 & 0.103 & 53.58 & \textbf{0.518} & 40s \\
            \textbf{Ours} & \textbf{0.042} & \textbf{78.67} & \textbf{0.060} & \textbf{67.46} & 0.492 & \textbf{30s} \\
        \bottomrule
        \bottomrule
      \end{tabular}
    }
  \end{subtable}
  \vfill
  \begin{subtable}[t]{0.45\linewidth}
  \caption{Results on Gen3DSR testset.}
    \label{tab:g3d_res} 
    \resizebox{\linewidth}{!}{
      \begin{tabular}{ccc}
        \toprule
        \toprule
           Method & CD$_S$$\downarrow$ & FS$_S$$\uparrow$ \\
        \midrule
        \midrule
            InstPIFu~\cite{liu2022instpifu} & 0.124 & 74.14 \\
            Gen3DSR~\cite{ardelean2025gen3dsr} & 0.120 & 68.82 \\
            MIDI~\cite{huang2025midi} & 0.566 & 23.49 \\
            \textbf{Ours} & \textbf{0.103} & \textbf{77.95} \\
        \bottomrule
        \bottomrule
      \end{tabular}
    }
  \end{subtable}
  \hfill
  \begin{subtable}[t]{0.53\linewidth}
  \caption{Results on ScanNet subset.}
    \label{tab:scan_res} 
    \resizebox{\linewidth}{!}{
      \begin{tabular}{ccc}
        \toprule
        \toprule
           Method & CD$_S$$\downarrow$ & FS$_S$$\uparrow$ \\
        \midrule
        \midrule
            Gen3DSR~\cite{ardelean2025gen3dsr} & 0.167 & 50.19 \\
            MIDI~\cite{huang2025midi} & 0.163 & 37.10 \\
            \textbf{Ours} & \textbf{0.130} & \textbf{61.58} \\
        \bottomrule
        \bottomrule
      \end{tabular}
    }
  \end{subtable}
  \vfill
  \begin{subtable}[t]{\linewidth}
  \caption{Results on our testset.}
    \label{tab:ours_res} 
    \resizebox{\linewidth}{!}{
      \begin{tabular}{cccccc}
        \toprule
        \toprule
           Method & CD$_S$$\downarrow$ & FS$_S$$\uparrow$ & CD$_O$$\downarrow$ & FS$_O$$\uparrow$ & IoU$\uparrow$ \\
        \midrule
        \midrule
            Gen3DSR & 0.265 & 46.72 & 0.546 & 31.95 & 0.304 \\
            MIDI & 0.801 & 15.35 & \textbf{0.179} & 35.99 & 0.033 \\
            Ours & \textbf{0.159} & \textbf{68.82} & 0.197 & \textbf{57.85} & \textbf{0.519} \\
        \bottomrule
        \bottomrule
      \end{tabular}
    }
  \end{subtable}
  \vspace{-13pt}
\end{table}

\begin{table}[h!]
    \centering
    % 调整列间距，去掉竖线后可以稍微宽一点点
    \setlength{\tabcolsep}{2.5pt}
    
    \caption{Ablation studies. We evaluate the number of layers ($\#K$), the coarse-to-fine strategy (C2F), the use of alignment loss (AL),
            the use of depth ratio embeddings (Dpt.), the inclusion of global features (S.), and the mixture of estimated geometry source from
            multiple geometry estimation methods (S. Dpt. v.s. M. Dpt.).}\label{tab:ablation}
    
    \resizebox{0.95\columnwidth}{!}{%
        \begin{tabular}{c|c|c|ccccc}
            \toprule
            \toprule
           $\#K$ & C2F & AL & CD$_S$$\downarrow$ & FS$_S$$\uparrow$ & CD$_O$$\downarrow$ & FS$_O$$\uparrow$ & IoU$\uparrow$ \\
           \midrule
           $12$ & \XSolidBrush & \XSolidBrush & 0.276 & 53.87 & \textbf{0.271} & \textbf{52.28} & 44.00 \\
           $12$ & \Checkmark & \XSolidBrush & 0.266 & 54.44 & 0.329 & 50.28 & 44.52 \\
           $12$ & \Checkmark & \Checkmark & \textbf{0.264} & \textbf{54.55} & 0.283 & 51.63 & \textbf{45.73} \\
           \midrule
           $6$ & \Checkmark & \XSolidBrush & 0.267 & 53.52 & 0.315 & 48.51 & 42.91 \\
           $12$ & \Checkmark & \XSolidBrush & 0.266 & 54.44 & 0.329 & 50.28 & 44.52 \\
           $18$ & \Checkmark & \XSolidBrush & \textbf{0.252} & \textbf{56.40} & \textbf{0.273} & \textbf{52.29} & \textbf{47.32} \\
           \midrule
           \midrule
           $\#K$ & Dpt. & S. & CD$_S$$\downarrow$ & FS$_S$$\uparrow$ & CD$_O$$\downarrow$ & FS$_O$$\uparrow$ & IoU$\uparrow$ \\
           \midrule
           $12$ & \XSolidBrush & \XSolidBrush & 0.276 & 50.53 & 0.435 & 43.10 & 36.93 \\
           $12$ & \Checkmark & \XSolidBrush & \textbf{0.259} & 50.85 & 0.368 & 43.37 & 37.90 \\
           $12$ & \XSolidBrush & \Checkmark & 0.274 & 50.60 & 0.440 & 42.77 & 36.24 \\
           $12$ & \Checkmark & \Checkmark & 0.266 & \textbf{54.44} & \textbf{0.329} & \textbf{50.28} & \textbf{44.52} \\
           \midrule
           \midrule
           $\#K$ & S. Dpt. & M. Dpt. & CD$_S$$\downarrow$ & FS$_S$$\uparrow$ & CD$_O$$\downarrow$ & FS$_O$$\uparrow$ & IoU$\uparrow$ \\
           \midrule
           $12$ & \Checkmark &  & \textbf{0.266} & 54.44 & 0.329 & 50.28 & 44.52 \\
           $12$ &  & \Checkmark & 0.272 & \textbf{55.20} & \textbf{0.262} & \textbf{53.22} & \textbf{47.33} \\
           \bottomrule       
           \bottomrule       
        \end{tabular}%
}
\vspace{-10pt}
\end{table}

\begin{figure*}
  \vspace{-6pt}
  \centering
  \begin{subfigure}{0.95\linewidth}
    \begin{overpic}[width=\linewidth]{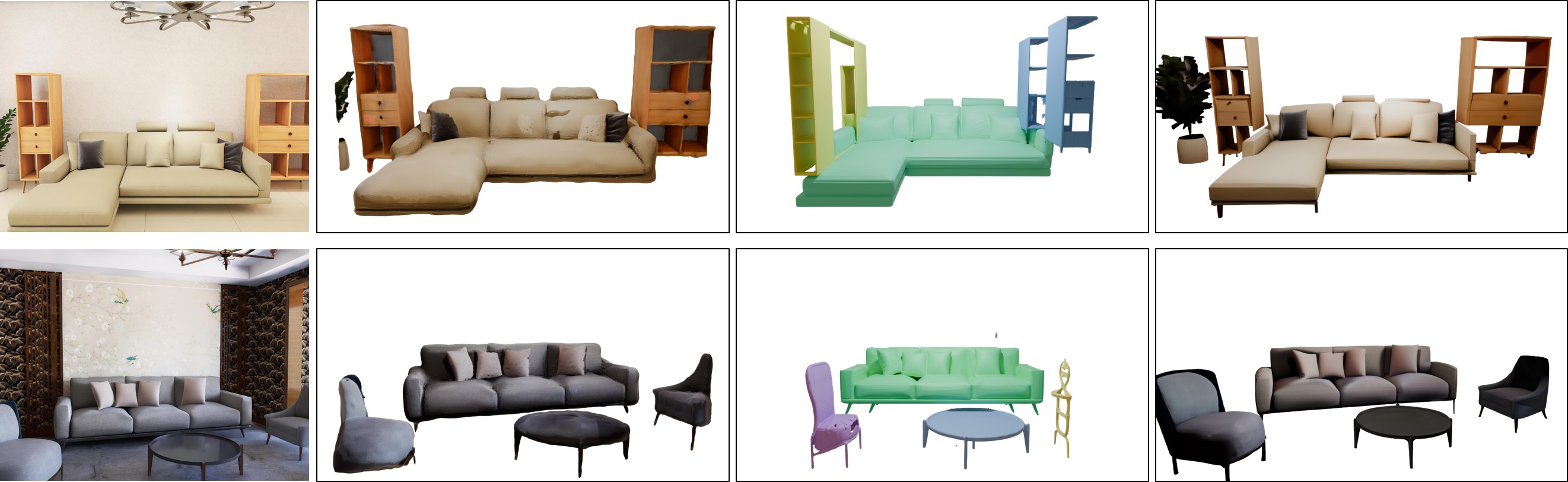}\small
    \put(5, 28){Input image}
    \put(30, 28){Gen3DSR}
    \put(58, 28){MIDI}
    \put(85, 28){Ours}
    \end{overpic}
    % \vspace{2pt}
    \caption{Visual comparisons on Gen3DSR testset.}
    % \vspace{2pt}
    \label{fig:gen3dsr_comp}
  \end{subfigure}
  \begin{subfigure}{0.95\linewidth}
    \begin{overpic}[width=\linewidth]{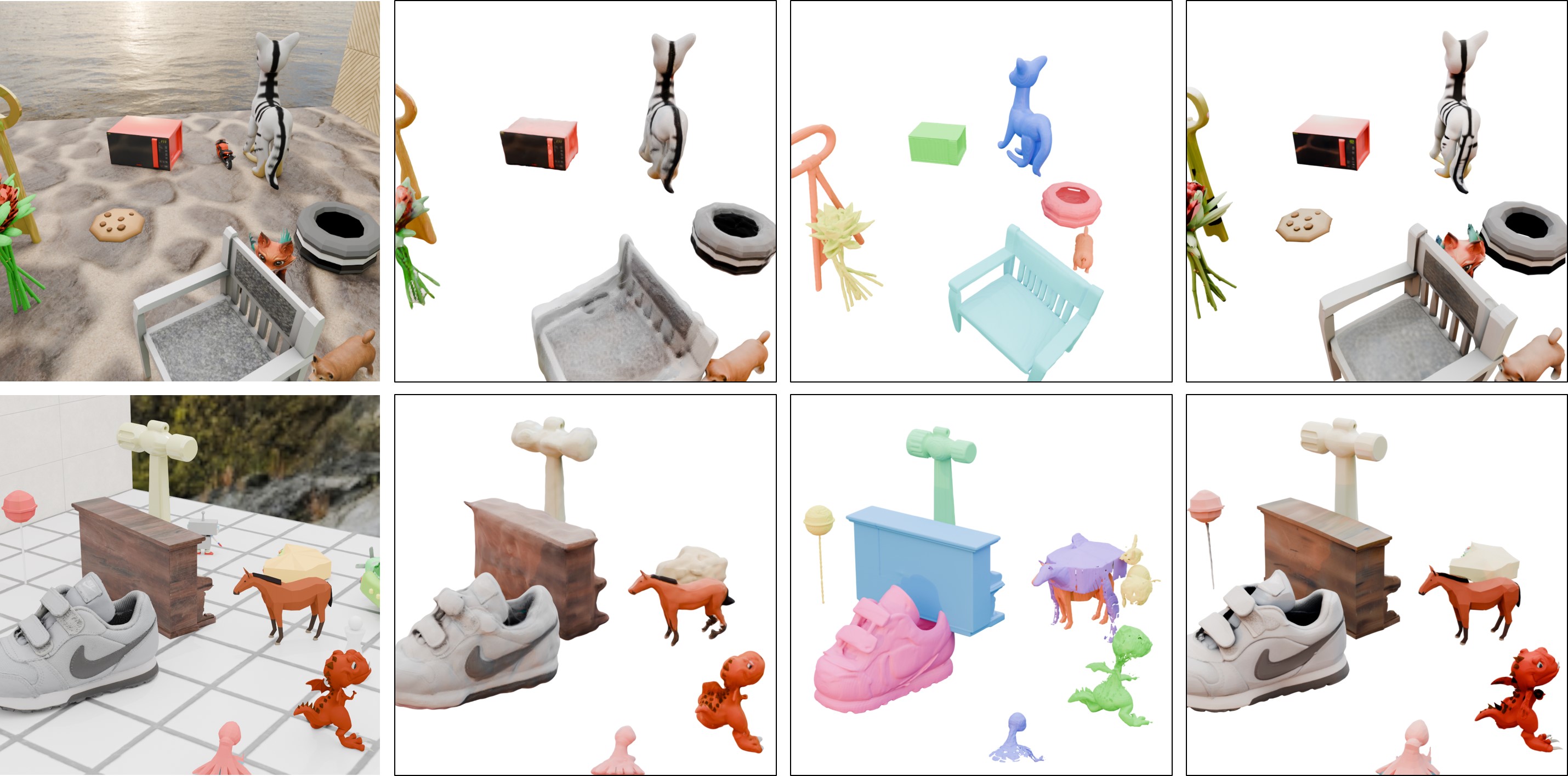}\small
    \put(7.5, 47){Input image}
    \put(34, 47){Gen3DSR}
    \put(60.5, 47){MIDI}
    \put(85, 47){Ours}
    \end{overpic}
    % \vspace{2pt}
    \caption{Visual comparisons on our testset.}
    % \vspace{2pt}
    \label{fig:our_test_comp}
  \end{subfigure}
  \vspace{-6pt}
  \caption{Visualization of the results on the Gen3DSR testset and our testset. The results on the Gen3DSR testset demonstrate the robustness of our scheme
  across different scenes, while the results on our test set show the great potential of our scheme in handling complex scenes.}
  \label{fig:comp_syn}
  \vspace{-10pt}
\end{figure*}
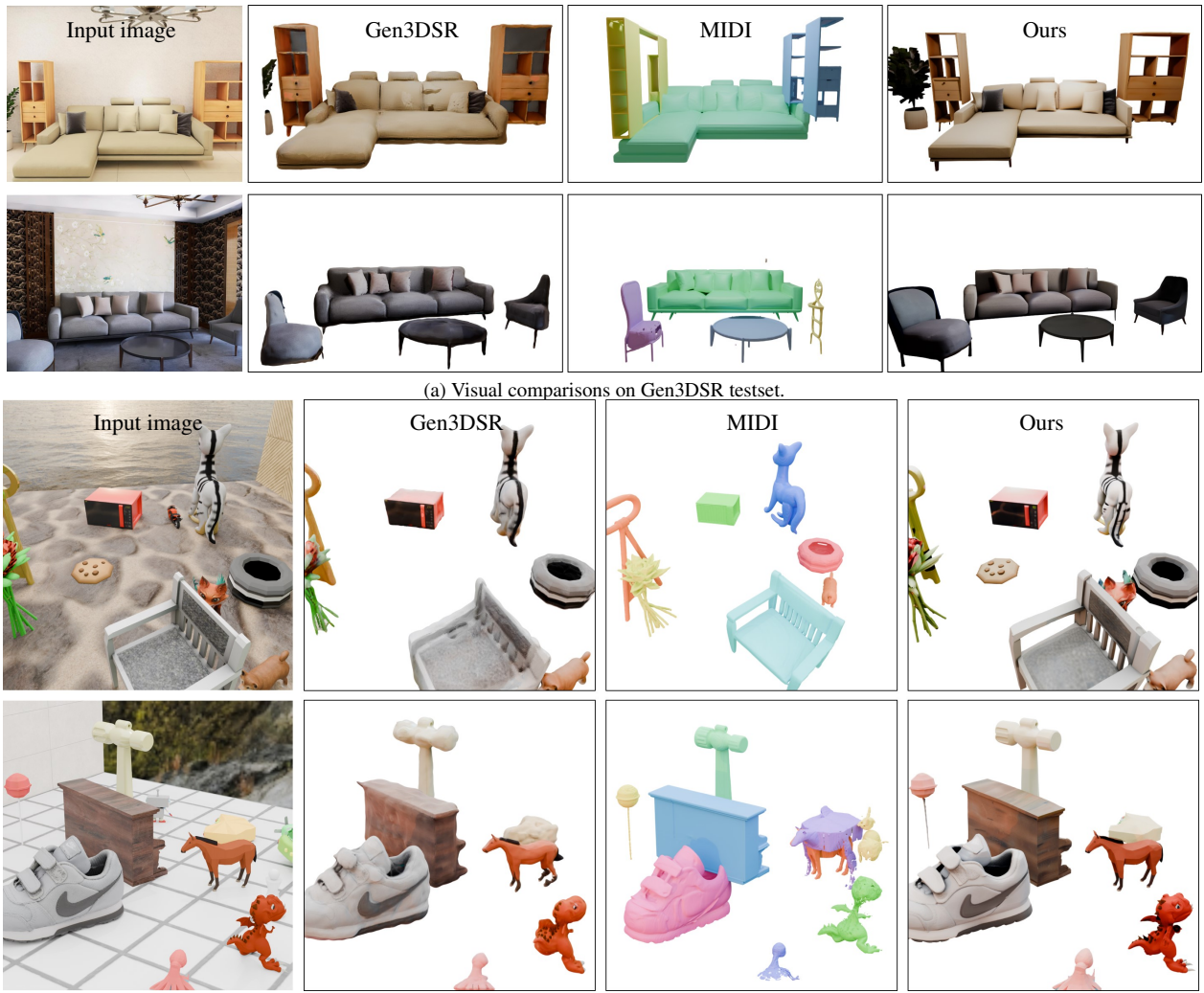

\subsection{Results on Synthetic Dataset}
\label{subsec:syn_eval}

As shown in \tabref{tab:midi_res} and \tabref{tab:g3d_res}, we report the quantitative metrics on the 3D-Front testset from MIDI~\cite{huang2025midi}
and the testset from Gen3DSR~\cite{ardelean2025gen3dsr}.
Our 3D-Fixer, only trained on our proposed dataset, significantly outperforms the existing methods,
which demonstrates the success of our scheme and the value of our ARSG-110K dataset.
Instead of relying solely on cross-attention to learn the scene-level spatial knowledge and object generation priors,
3D-Fixer flexibly combines the spatial knowledge from geometry estimation priors 
with the object-level priors from 3D generation models, 
fully leveraging the strengths of both methods and achieving robust yet superior performance.
The object-level metrics demonstrate that our 3D-Fixer fully utilizes the knowledge from 3D generative priors
to generate high-quality and scene-aligned 3D assets.
Moreover, the scene-level metrics further demonstrate the effectiveness of integrating geometry estimation priors 
in scene generation, which has been underestimated before.
Most significantly, our scheme exhibits robustness and generalization across different scenes, 
which benefits from the in-place completion scheme that effectively exploits the layout priors from geometry estimation
and injects the layout knowledge into 3D generation priors.

As shown in \figref{fig:gen3dsr_comp}, the feed-forward diffusion method MIDI suffers from the out-of-domain problem, as these scenes are more complex with varying numbers of furniture.
Meanwhile, per-instance generation and optimization method Gen3DSR is more robust, but the generation quality is lower than our scheme.
On the contrary, our scheme generates high-quality 3D assets and accurately captures the spatial relationship between each instance, which further proves the effectiveness of 3D-Fixer.

\subsection{Results on Complex Dataset}
\label{subsec:real_eval}

\myPara{Real-world dataset}
As reported in \tabref{tab:scan_res}, we evaluate related methods on 15 views from the ScanNet dataset~\cite{dai2017scannet}, and use MetaScenes~\cite{yu2025metascenes} as the ground truth.
The metrics demonstrate the robustness and generalization ability of our scheme to real-world scenes.
For visual comparisons and the evaluation scenes, please refer to the Supplementary.

\myPara{Synthetic dataset}
We evaluate related works on our proposed testset in \tabref{tab:ours_res}.
To ensure a fair comparison, we select, for each scene, the five instances with the 
highest pixel-validation ratios, given that MIDI is trained on a small-scale indoor dataset.
The scene-level metrics demonstrate that MIDI suffers from the out-of-domain problem in complex scenes.
Gen3DSR exhibits robustness, but the way it utilizes 3D generative priors restricts its performance.
In contrast, our scheme demonstrates robustness and efficiency.

For visual comparisons, we evaluate all methods using all instance masks, as shown in \figref{fig:our_test_comp}. With the increase in the number of instances, the cross-attention mechanism in MIDI struggles to capture scene-level interrelationships. Similarly, results from Gen3DSR \cite{ardelean2025gen3dsr} demonstrate a failure to fully exploit the potential of 3D generation priors, yielding lower-quality geometry. In contrast, our 3D-Fixer generates high-fidelity 3D assets while accurately capturing the spatial relationships among instances.

\subsection{Ablation Study}
\label{subsec:abl_exp}
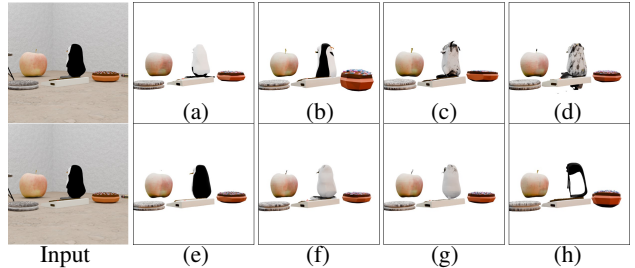
\begin{figure}[h]
  \centering
  \begin{subfigure}{0.99\linewidth}
  \begin{overpic}[width=\linewidth]{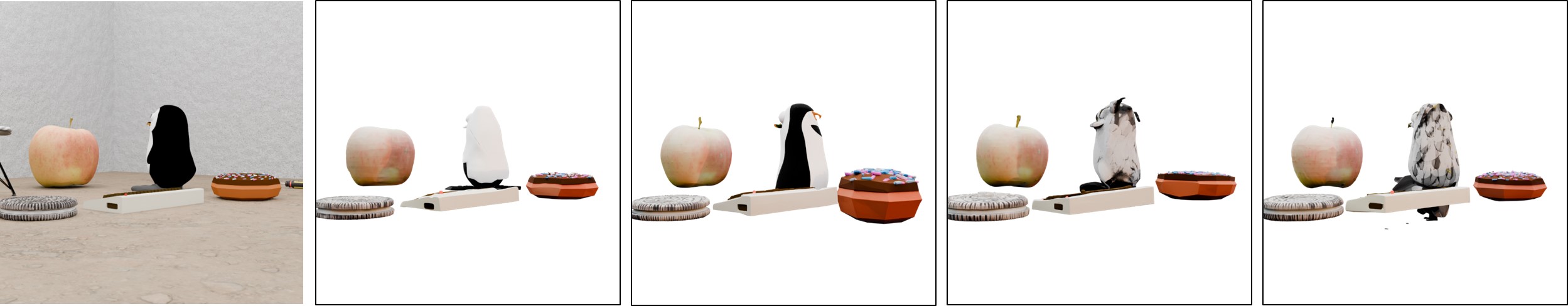}\small
    \put(28, 0.8){(a)}
    \put(47.9, 0.8){(b)}
    \put(68, 0.8){(c)}
    \put(87.8, 0.8){(d)}

    \end{overpic}
  \end{subfigure}
  \vspace{2pt}
  \begin{subfigure}{0.99\linewidth}
  \begin{overpic}[width=\linewidth]{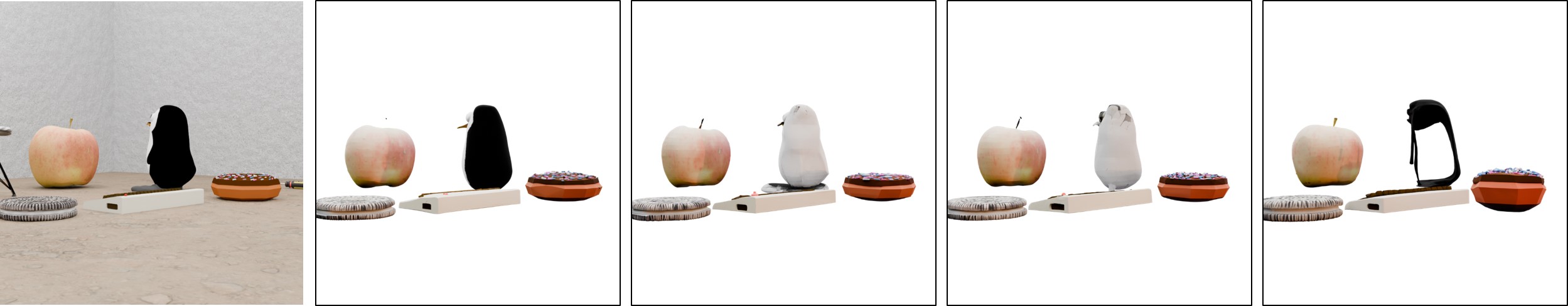}\small
    \put(5, -2.8){Input}
    \put(28, -2.8){(e)}
    \put(47.9, -2.8){(f)}
    \put(68, -2.8){(g)}
    \put(87.8, -2.8){(h)}

    \end{overpic}
  \end{subfigure}
  \caption{Visualization of ablation studies. Experiments (a)-(d) are designed to evaluate the coarse-to-fine (C2F) strategy and the network layers (K), which are as follows: (a) w/o C2F, K=12; (b) w/ C2F, K=6; (c) w/ C2F, K=12; w/ C2F, K=18. 
  Experiments (e)-(h) are designed to evaluate the Alignment Loss (AL),
  depth ratio embedding (Dpt.), and the global feature input (Glob.), which are as follows: (e) w/ C2F, K=12, AL, Dpt., and Glob.; 
  (f) w/o AL and Dpt.; (g) w/o AL and Glob.; (h) w/o AL, Dpt. and Glob.}
  
  \label{fig:res_abla}
  \vspace{-10pt}
\end{figure}

We conduct comprehensive ablation studies to analyze the effectiveness of our designs,
as in \tabref{tab:ablation} and \figref{fig:res_abla}.

\myPara{Training Strategy}
(a) Coarse-to-fine strategy. The coarse-to-fine strategy is designed to address scale uncertainty in scenes. The scene-level metrics demonstrate that this strategy significantly impacts the completeness of instances, thereby improving overall scene-level performance.

(b) Alignment loss. We introduce the alignment loss to constrain the 3D priors under occlusion. As depicted in the metrics, the alignment loss helps the model converge better. The visual results also confirm that, with the alignment loss, our method achieves superior visual quality.

(c) Mixture of estimated geometry sources. We also introduce multiple geometry estimation methods into our training data. This improves the robustness of our method by helping the model learn how to handle diverse perturbations, allowing it to better recover accurate geometry.

\myPara{Network Design}
(a) Number of layers in the network. We experiment with 6, 12, and 18 layers in 3D-Fixer to investigate the network design. The metrics show that deeper models converge to better numerical results. However, visualizations in \figref{fig:res_abla} indicate that the generated geometry is already complete when the number of layers reaches 12. Therefore, we use a 12-layer configuration for 3D-Fixer.

(b) Depth ratio embedding and global feature injection. We remove the depth ratio embedding and the global feature cross-attention to assess their impact. As shown in \tabref{tab:ablation} and \figref{fig:res_abla}, without either or both of these components, the model fails to generate complete geometry.

%% file: sec/6_conclusion.tex
\section{Conclusion}
\label{sec:con}

We present 3D-Fixer, a generalizable framework for single-image scene generation that synergizes geometric estimation with generative priors. By employing a novel in-place completion paradigm, our method eliminates error-prone alignment, achieving state-of-the-art fidelity and efficiency. Furthermore, we introduce ARSG-110K, the largest high-quality scene generation dataset to date, which we believe will serve as a foundational benchmark for future research.

\section*{Acknowledgment}
This work was supported by the National Key R\&D Program of China No. 2024YFC3015801, National Science Fund of China under Grant Nos. U24A20330, 62361166670, and 62276144.

%% file: sec/X_suppl.tex
\clearpage
\setcounter{page}{1}
\maketitlesupplementary

\section{Implementation details}
\label{sec:impl}

\myPara{Base model}
Our method builds on the TRELLIS~\cite{xiang2025structured} framework,
which is a two-stage rectified flow generation method, where the first stage is a DiT~\cite{peebles2023scalable} model 
that generates the sparse voxel structure in the latent space,
and the second stage is a DiT-style model based on the sparse coordinates predicted in stage one to generate the 
SLAT representation~\cite{xiang2025structured}.
The first stage utilizes 3D VAE to compress the $64^3$ volumetric grid with binary occupancy into 
a low-resolution feature grid at a resolution of 16 and each grid contains a vector of dimension 8.
The SLAT representation in TRELLIS is sparse volumetric feature representation,
which is a sparse voxel grid at a resolution of 64 and each activated voxel stores an 8-dimensional feature vector.
To construct this representation, TRELLIS first voxelizes a 3D asset into a sparse grid at resolution $64^3$.
It then renders dense views around the 3D asset and extracts per-view features via DINOv2 model~\cite{oquab2023dinov2}.
These image features are subsequently projected onto the corresponding sparse voxels.
Finally, a sparse 3D VAE encodes the aggregated DINOv2 features within each voxel into a compact 8-dimensional latent feature, producing the final SLAT representation.

\myPara{3D-Fixer details}
Our complete 3D-Fixer framework consists of three modules: the Coarse Structure Completer, the Fine Shape Refiner, 
and the Occlusion-Aware 3D Texturer.
The three components are constructed based on image-conditioned TRELLIS model.

The Coarse Structure Completer and the Fine Shape Refiner are designed to generate the voxel grids,
which are built on the first stage of TRELLIS model and each consists of 12 layers of our basic block as in Fig. 3 of the main paper.
During training, we randomly sample an estimated depth map $d_{\text{est}}$ from one of MoGe v2~\cite{wang2025moge}, VGGT~\cite{wang2025vggt}, DepthAnything v2~\cite{yang2024depth},
or Depth Pro~\cite{bochkovskii2024depth}.
The sampled depth map is mixed with the ground truth depth map $d_{\text{gt}}$ using a coefficient $\alpha$
as $d=\alpha\cdot d_{\text{est}} + (1-\alpha)\cdot d_{\text{gt}}$,
where $\alpha$ is uniformly sampled in $[0.0, 1.0]$.
The $\alpha$ is further encoded as a depth-ratio embedding and provided to the model;
during inference, we set $\alpha$ to $1.0$. 
The visible point cloud is voxelized into a $64^3$ volumetric grid, 
encoded into the latent space via the pre-trained 3D VAE from TRELLIS,
and then supplied to both the Coarse Structure Completer and the Fine Shape Refiner as the partial geometry features.
For the global geometry conditioning, we use the MoGe v2~\cite{wang2025moge} to extrach feature tokens from the scene image,
while the occluded conditioning is provided by instance-level image tokens from DINOv2.
The Occlusion-Aware 3D Texturer generates textured 3D assets conditioned on the voxels produced by the first stage,
which is built on the second-stage TRELLIS architecture and similarly comprises 12 layers of our basic block.
To provide 3D-aware texture cues, we project the scene-level global features onto the voxel grid.
Additionally, we calculate the visibility ratio of the voxel grid with respect to the input view
and encode this value as a visibility ratio embedding,
which is also supplied to the model.
The global geometry conditioning and occlusion-aware conditioning follow the same design as in our first-stage models.

\myPara{Training}
We train the 3D-Fixer on our proposed dataset.
Because 3D instances in the scenes are randomly placed, we first fine-tune the base models using randomly rotated 3D assets to enhance the priors. The first-stage model is fine-tuned on 32 NVIDIA RTX 5090 GPUs for 150K steps with a batch size of 128. The second-stage model is fine-tuned on 32 NVIDIA RTX 5090 GPUs for 450K steps with a batch size of 128. We also fine-tune the mesh decoder and the 3D Gaussian Splatting decoder on 32 NVIDIA RTX 5090 GPUs for 80K steps with a batch size of 128. For all fine-tuning stages, we use the AdamW~\cite{loshchilov2017decoupled} optimizer with a learning rate of $1e-5$. The Coarse Structure Completer and the Fine Shape Refiner are trained separately on 32 NVIDIA RTX 5090 GPUs for 80K steps with a batch size of 128. Before training on the scene-level dataset, we first pre-train the models on an object-level dataset for 100K steps. The Occlusion-Aware 3D Texturer is trained separately on 32 NVIDIA RTX 5090 GPUs for 90K steps with a batch size of 128 on the scene-level dataset.
The Coarse Structure Completer and the Fine Shape Refiner are separately trained on 
32 NVIDIA RTX 5090 GPUs for 80K steps with a batch size of 128.
Before training on the scene-level dataset, we first pre-train the models on object-level dataset for 100K steps.
The Occlusion-Aware 3D Texturer is trained separately on 32 NVIDIA RTX 5090 GPUs for 90K steps with a batch size of 128 on scene-level dataset.
We use AdamW~\cite{loshchilov2017decoupled} optimizer with a learning rate of $5e-5$.
In addition to the standard flow matching loss, we apply our proposed alignment loss 
to the three models with weighting factors of 0.1, 0.5, and 0.5, respectively.
During inference, we use the classifier-free guidance with a guidance strength of 5,
and the sampling steps are set to 25.

\section{Robustness to input noise}
\label{sec:robust}

\begin{figure}
  \vspace{-6pt}
  \centering
  \begin{subfigure}{0.95\linewidth}
    \begin{overpic}[width=\linewidth]{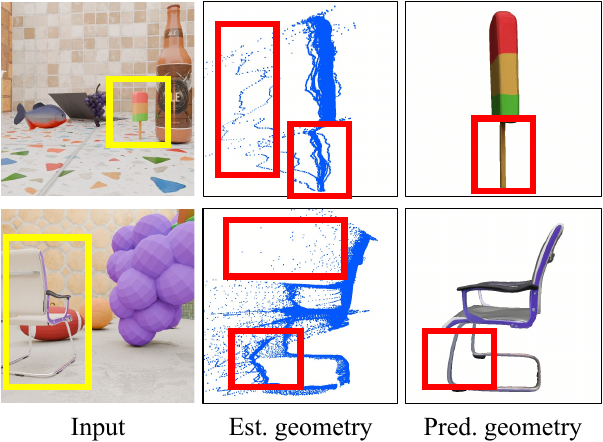}
    \end{overpic}
    % \vspace{2pt}
    \caption{Handle geometry distortion.}
    % \vspace{2pt}
    \label{fig:init_dist}
  \end{subfigure}
  \begin{subfigure}{0.95\linewidth}
    \begin{overpic}[width=\linewidth]{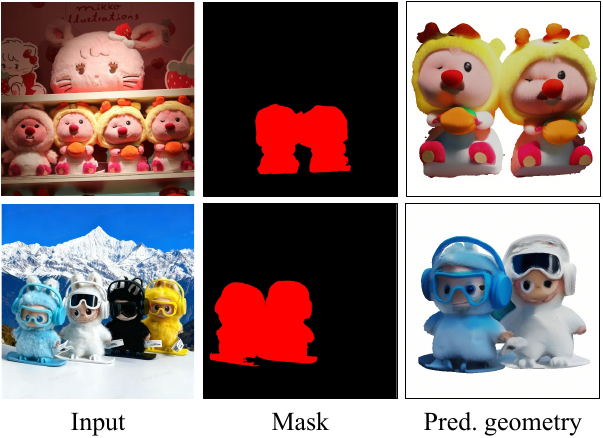}
    \end{overpic}
    \caption{Handle mask error.}
    \label{fig:mask_error}
  \end{subfigure}
  \vspace{-6pt}
  \caption{Visualization of our scheme handling initial geometry distortions and mask errors.}
  \vspace{-10pt}
\end{figure}

\begin{figure*}
  \vspace{-6pt}
  \centering
  \begin{subfigure}{0.95\linewidth}
    \begin{overpic}[width=\linewidth]{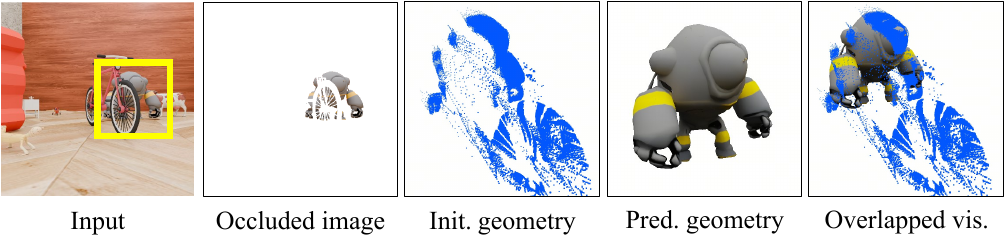}
    \end{overpic}
  \end{subfigure}
  \vspace{-6pt}
  \caption{Visualization of our method handling complex occlusion patterns.}
  \label{fig:comp_occ}
  \vspace{-10pt}
\end{figure*}

\myPara{Tolerance to initial distortion}
Although the initial geometries are distorted, 
our ARSG-110K contains massive high-quality 3D GT as supervision samples,
therefore, 3D-Fixer can learn how to generate plausible 3D assets.
Furthermore, we introduce three designs to improve the model's ability.
First, the ORFA strategy (Sec. 3.5 in the main paper) provides detailed supervision.
Second, the mixed source of initial geometries (\secref{sec:impl}) augments the supervision samples with more diverse distortion patterns.
Third, the depth-ratio embedding (\secref{sec:impl}) helps 3D-Fixer to handle distortions of varying degrees.
To quantify the robustness, we linearly interpolate the predicted depth and GT depth with a coefficient $\alpha$ to mimic different distortions.
We report object-level ($O$) Chamfer Distance (CD), F1-Score@0.1 (FS), and Bounding Box IoU in \tabref{tab:dist_res}.
The metrics and \figref{fig:init_dist} indicate 3D-Fixer's robustness to initial distortions.

\myPara{Tolerance to mask errors}
Our scheme is able to handle mask errors, where multiple instances are merged into one mask as shown in \tabref{fig:mask_error}.

\myPara{Robustness to complex occlusion patterns}
Complex occlusion patterns cause severe distortions to the initial geometries,
but our dataset and Contextual Conditioning design (Sec. 3.3 in the main paper) enable our method to robustly handle complex occlusion.
First, our dataset contains complex occlusion patterns.
On our testset,
40.18\% of instance masks have more than one 8-connected component,
and 10.89\% have more than four.
Second, our module jointly processes fragmented geometry and global features 
as mentioned in Sec. 3.3 of the main paper, which enables 3D-Fixer to reason about relationships among multiple visible parts.
As shown in \figref{fig:comp_occ}, 3D-Fixer can extract reliable information from the fragmented geometry rather than fully trusting it.

\section{Texture quality}
\label{sec:tex}

To access the quality of the synthesized texture, we separately render three views for the visible ($V$) and unseen ($U$) region on our testset, 
and report FID and CLIP score in \tabref{tab:rend_res}.
These metrics demonstrate the quality and semantic consistency of the generated textures.

\section{More results}
\label{sec:more_res}

\begin{table}[t]
  \centering
  \vspace{-5pt}
  \caption{Quantitative comparisons on our testset. We report Object ($O$) level Chamfer Distance (CD) and F-Score (FS), 
  along with Bounding Box IoU, to quantify the robustness of our scheme on geometry distortion.
  We also report the object-level rendering metrics on our testset to quantify the texture quality compared to Gen3DSR (G3D).}
  \vspace{-8pt}
  \begin{subtable}[t]{0.35\linewidth}
  \caption{Distortion robustness.}
    \label{tab:dist_res} 
    \resizebox{\linewidth}{!}{
      \begin{tabular}{cccc}
        \toprule
        \toprule
           $\alpha$ & CD$_O$$\downarrow$ & FS$_O$$\uparrow$ & IoU$\uparrow$ \\
        \midrule
        \midrule
            0.2 & \textbf{0.185} & \textbf{60.55} & \textbf{0.547} \\
            0.6 & 0.188 & 58.95 & 0.528 \\
            1.0 & 0.197 & 57.85 & 0.519 \\
        \bottomrule
        \bottomrule
      \end{tabular}
    }
  \end{subtable}
  \hfill
  \begin{subtable}[t]{0.62\linewidth}
  \caption{Rendering metrics.}
    \label{tab:rend_res} 
    \resizebox{\linewidth}{!}{
      \begin{tabular}{ccc}
        \toprule
        \toprule
           Method & FID$_V$/FID$_U$$\downarrow$ & CLIP$_V$/CLIP$_U$$\uparrow$ \\
        \midrule
        \midrule
            G3D~\cite{ardelean2025gen3dsr} & 102.72 / 119.22 & 80.70 / 77.71 \\
            \textbf{Ours} & \textbf{43.52 / 46.25} & \textbf{89.82 / 88.51} \\
        \bottomrule
        \bottomrule
      \end{tabular}
    }
  \end{subtable}
  \vspace{-13pt}
\end{table}

\begin{figure*}
  \centering
  \begin{subfigure}{0.95\linewidth}
    \begin{overpic}[width=\linewidth]{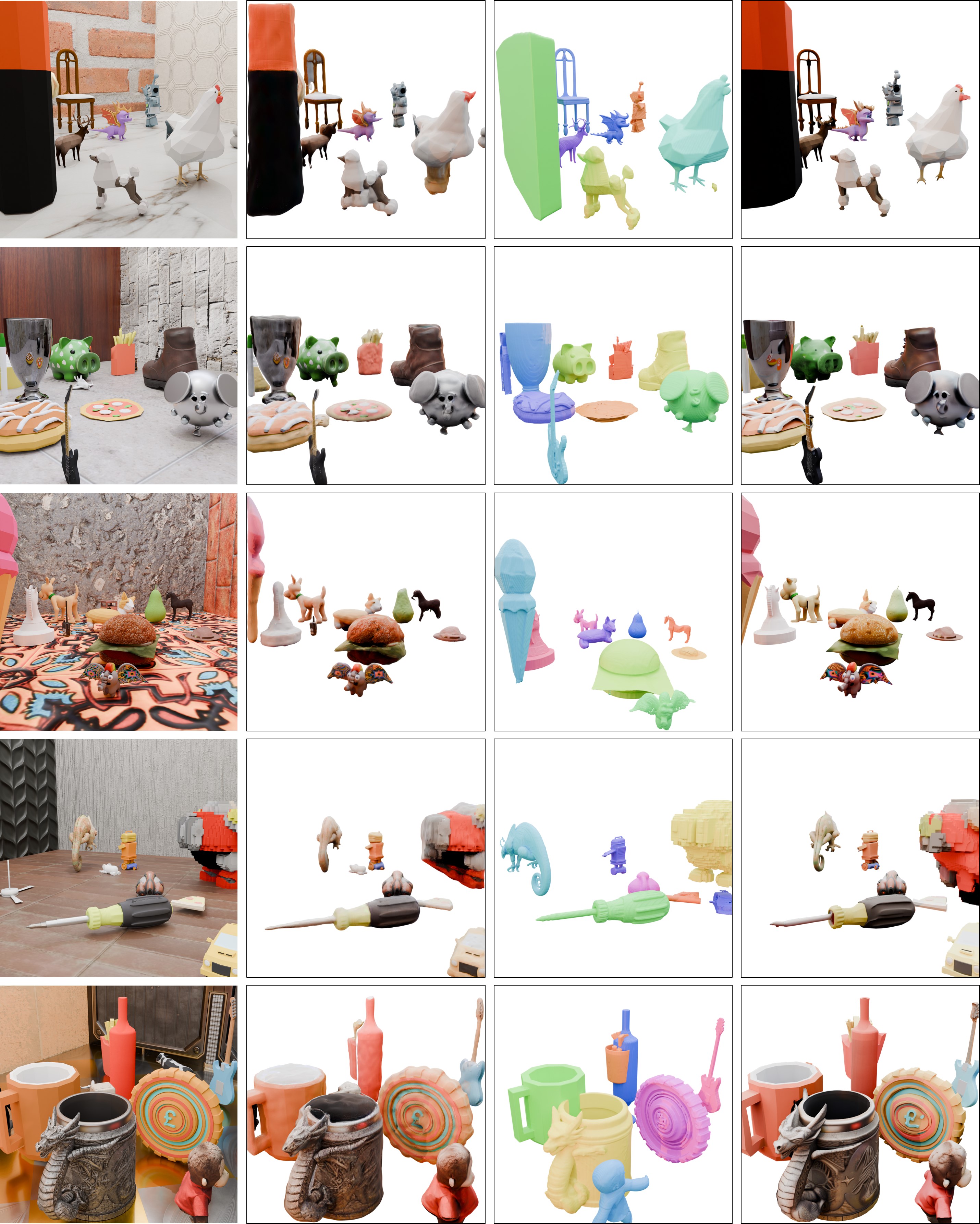}\small
    \put(6, -2){Input image}
    \put(26.8, -2){Gen3DSR}
    \put(48, -2){MIDI}
    \put(69, -2){Ours}
    \end{overpic}
  \end{subfigure}
  \vspace{4pt}
  \caption{Visual comparisons on our testset.}
  \label{fig:our_test_comp_supp}
\end{figure*}

In this section and in our Supplementary Video, we present diverse visualizations across a variety of scenarios.
As in \figref{fig:our_test_comp_supp}, our approach produces high-quality 3D assets 
and accurately reconstructs the spatial layout of the scene.
In contrast, Gen3DSR generates blurry geometric structures, while MIDI fails to recover an accurate spatial layout.

\begin{figure*}
  \centering
  \begin{subfigure}{0.95\linewidth}
    \begin{overpic}[width=\linewidth]{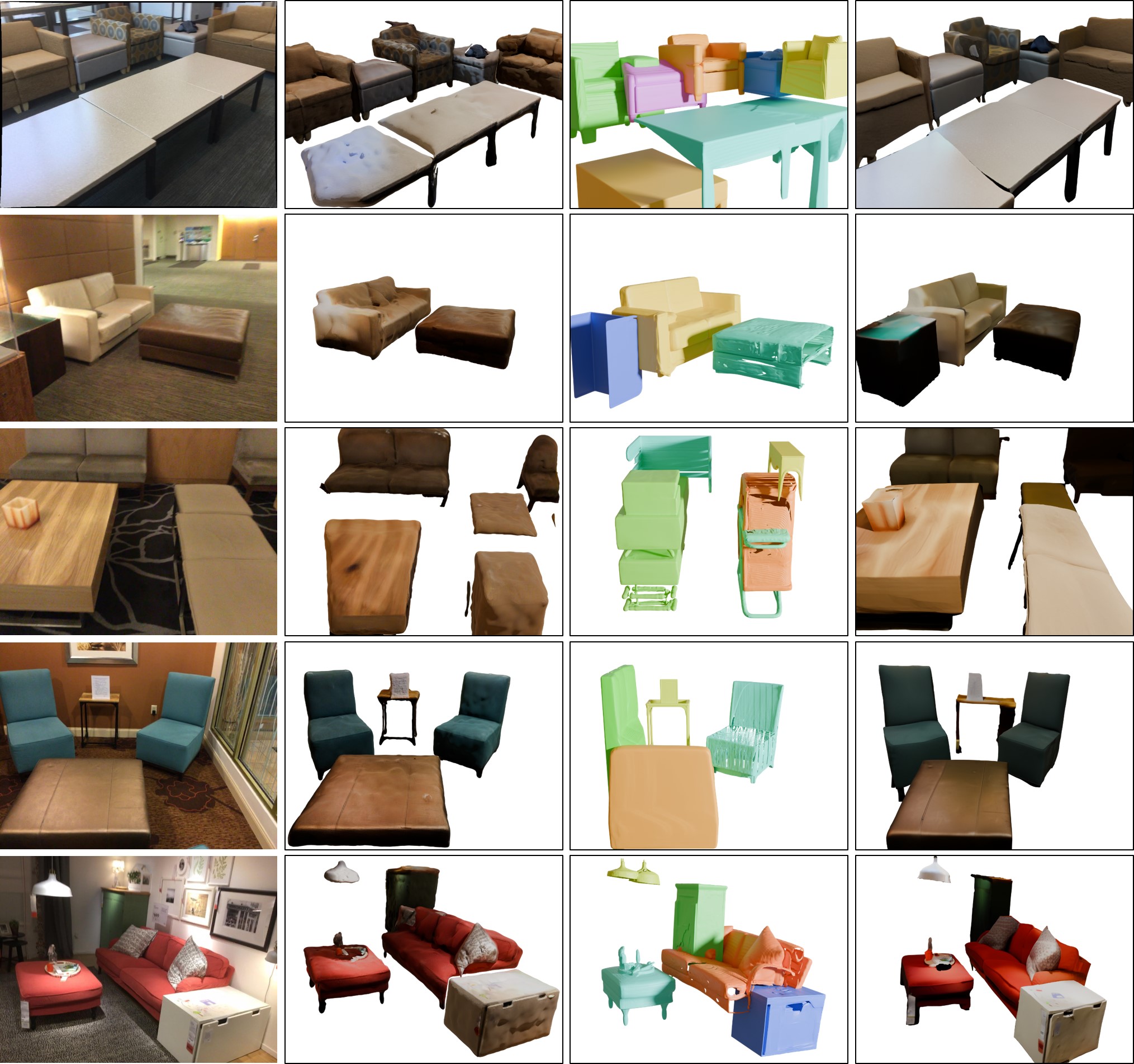}\small
    \put(8, -2){Input image}
    \put(34, -2){Gen3DSR}
    \put(60, -2){MIDI}
    \put(86, -2){Ours}
    \end{overpic}
  \end{subfigure}
  \vspace{4pt}
  \caption{Visual comparisons on ScanNet.}
  \label{fig:real_scanet_supp}
\end{figure*}

As in \figref{fig:real_scanet_supp}, we further evaluate our method on real-world indoor scenes from ScanNet~\cite{dai2017scannet}.
Our approach generalizes well to real scenes and produces coherent and high-quality 3D assets with accurate layout.
However, Gen3DSR yields fragmented geometry and MIDI struggles to generate the accurate spatial layout.
Furthermore, We report the real-world performance in Tab. 2c in the main paper on ScanNet dataset,
where we use the following subset for evaluation: 
frame 360 from scene0048\_00, frame 680 from scene0036\_00, frame 105 from scene0033\_00, 
frame 160 from scene0031\_00, frame 440 from scene0028\_00, frame 248 from scene0053\_00, 
frame 235 from scene0087\_00, frame 210 from scene0081\_00, frame 105 from scene0199\_00, 
frame 420 from scene0160\_00, frame 200 from scene0162\_00, frame 60 from scene0165\_00, 
frame 1060 from scene0148\_00, frame 940 from scene0134\_00, and frame 1441 from scene0129\_00.

\begin{figure*}
  \centering
  \begin{subfigure}{0.95\linewidth}
    \begin{overpic}[width=\linewidth]{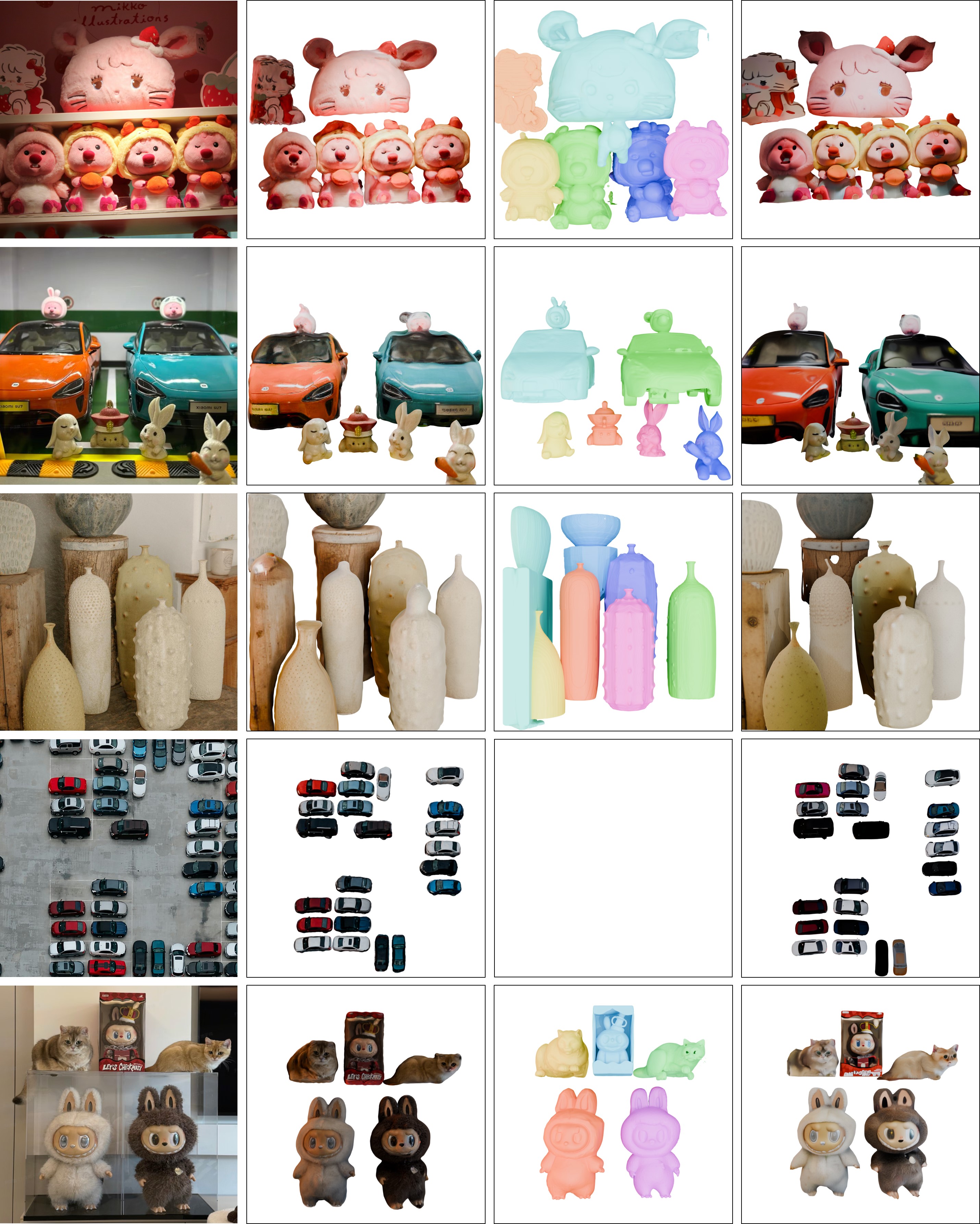}\small
    \put(45, 29.5){\textcolor{gray}{Out of Memory}}
    \put(6, -2){Input image}
    \put(26.8, -2){Gen3DSR}
    \put(48, -2){MIDI}
    \put(69, -2){Ours}
    \end{overpic}
  \end{subfigure}
  \vspace{4pt}
  \caption{Visual comparisons on real world captured images.}
  \label{fig:real_cap_supp}
\end{figure*}

As in \figref{fig:real_cap_supp}, we further evaluate our method on more challenging real captured scenes.
Even in scenarios with complex layouts or a large number of instances, our method successfully generates accurate spatial arrangements and geometry,
In contrast, MIDI encounters out-of-memory failures on an NVIDIA RTX 4090 GPU with 24 GB memory when dealing with scenes with large amounts of instances,
and Gen3DSR generates fragmented or low quality geometries.

\section{Procedural scene generation}
\label{sec:blender_scene}

To procedurally construct the ARSG-110K dataset, we use a subset of 180K high-quality 3D object assets from TRELLIS-500K~\cite{xiang2025structured}.
To improve rendering photorealism, we additionally collect over 1K HDR maps and 5K material textures from BlenderKit, a community platform for sharing 3D assets.
All scenes are rendered using the Blender Cycles engine.
For each scene, we first create a floor plane, and then probabilistically place 0 to 4 additional planes around it as walls to simulate both indoor and outdoor environments.
A material texture is randomly assigned to each plane.
We also randomly select an HDR map for scene illumination.
For object placement, we randomly sample 20 3D assets from the object pool, normalize each instance, apply a random rotation around the z-axis and a random scaling factor within $[0.5, 2.0]$, and then place the instances into the scene sequentially.
To avoid interpenetration, each object is placed with at most 100 attempts. In each attempt, random scaling and rotation are applied, followed by collision detection against previously placed objects. The placement process terminates when a collision-free configuration is found or the maximum number of attempts is reached.
The dataset and the scene construction script will be made publicly available.

\section{Limitations and Future Works}
\label{sec:limits}

As our method performs in-place completion using geometry-based cues for scene generation,
the accuracy of the recovered layout inherently depends on the quality of the initial estimated geometry.
We believe an important future direction is to explore unified frameworks to simultaneously estimate the scene geometry 
and generate the complete 3D instances in the scene.